# XALIGN: A Human-centric Visual Analytics Approach to Understanding AI Explanations

Bahavathy Kathirgamanathan[a,*], Gennady Andrienko[b,c], Natalia Andrienko[b,c], Tatiana von Landesberger[a]

[a] *University of Cologne, Cologne, Germany*
[b] *Fraunhofer IAIS, Sankt Augustin, Germany*
[c] *City St George's Institute of London, London, UK*

**Abstract**

Explainable AI (XAI) methods are usually evaluated for technical properties such as faithfulness and robustness. In high-stakes application domains, experts also need to check model predictions against established knowledge. Different explanation methods often disagree, and this disagreement is hard to see when each method is inspected separately. We propose XALIGN, a visual analytics workflow and interface that shows rule-based explanations and several feature attribution methods (TreeSHAP, KernelSHAP, and LIME) side by side for the same Random Forest model. Agreements and contradictions between the methods become visible signals that users can explore and compare with domain knowledge. We apply XALIGN in two case studies: the German Credit dataset and a meteorological dataset, in which events derived from ICON model output and Cloudnet observations are classified. In both, the views reveal discrepancies between the rule-based explanation and the attribution methods that a single method would not show. In an expert evaluation with three specialists, participants found XALIGN useful for identifying influential features and for avoiding over-trust in predictions. Linking rules to attributions was harder for the participant without an XAI background.

*Corresponding author
*Email address:* bkathirg@uni-koeln.de (Bahavathy Kathirgamanathan)

## 1. Introduction

Traditional explainable AI (XAI) methods are predominantly model-centric, focusing on describing how a model arrives at its predictions while often paying less attention to whether these explanations are meaningful, insightful, or useful to human users. Such methods are commonly evaluated against technical properties such as faithfulness, robustness, and computational efficiency [1]. Although these properties are important for assessing explanation quality, they do not fully capture the needs of human users. An explanation can be faithful to a flawed model, stable yet uninformative, or computationally efficient while providing limited value to the person interpreting it. We therefore argue that a purely model-centric approach to explanation is insufficient for developing AI systems that effectively support human decision-making, particularly in high-stakes domains where domain experts need to understand, interpret, and validate AI recommendations. This motivates a shift toward human-centred XAI, where explanations are designed to characterise model behaviour, to support human reasoning, and to incorporate relevant domain knowledge [2, 3].

It is well established that different explanation techniques do not necessarily provide consistent views of model behaviour [4], with each approach carrying its own strengths and limitations. Feature attribution methods such as SHAP [5] and LIME [6] are widely used due to their flexibility, model-agnostic nature, and applicability to high-dimensional data. However, the attributions they produce do not always correspond to human reasoning or domain expectations, and concerns about their stability and consistency have been well documented [7]. Rule-based methods, by contrast, describe model decisions through logical if-then relationships, offering explanations that are often more intuitive and closer to the reasoning styles of domain experts [8]. However, extracting and presenting meaningful rules becomes increasingly difficult as model and dataset complexity grows, limiting their scalability in some real-world applications. Hybrid approaches offer an opportunity to examine these perspectives together, highlighting agreements and discrepancies between explanation techniques and revealing cases where model behaviour diverges from domain expectations [4]. By combining complementary explanation strategies, such approaches can provide a richer basis for interpreting model behaviour and developing explanations that are better aligned with human knowledge.

In this work, we advocate for interactive visual analytics frameworks that enable users to explore multiple explanation techniques from a human-centered perspective. Interactive visual analytics can allow users to dynamically compare explanations, investigate feature interactions, and examine areas where different explanation methods provide conflicting interpretations. Such interactions can also enable domain experts to identify surprising, unexpected, or potentially illogical model behaviours and flag them for further investigation. In this context, visual analytics provides a means of bringing human knowledge into the interpretation and validation of model explanations, rather than treating explanations as static outputs of an XAI method. The resulting human-in-the-loop process can support iterative investigation, where insights from domain experts inform the interpretation of explanations and potentially contribute to subsequent model development. In this way, such a framework aims to bridge the gap between model-centric XAI and real-world decision-making by placing human understanding and domain knowledge at the center of the explanation process. Rather than viewing explanations simply as a mechanism for increasing confidence in AI, we focus on supporting users in developing an informed understanding of when and why model outputs should be examined, questioned, or incorporated into their decision-making.

We propose XALIGN, a visual analytics workflow and interface for comparing rule-based and feature attribution explanations of the same model. Users examine where the explanations agree, find regions where they conflict, and judge these regions against domain knowledge. The outcome can inform the refinement of data, rules, and the choice of explainer.

This paper extends our earlier workshop paper [9]. Compared with that paper, we (1) apply the approach in two case studies, on the German Credit dataset and on meteorological data, (2) extend the interaction in the interface, for example with a tab for side-by-side comparison of attribution methods and classes and some improved user interaction tools, and (3) report an expert evaluation with three specialists.

## 2. Related Work

### *2.1. Limitations of Model-Centric XAI*

Model-centric XAI typically reports which features influence a model's prediction, but What this frequently overlooks is whether the explanations are meaningful or actionable to the people who must use them. There is growing recognition that the usefulness of an explanation is not an intrinsic property of the method, but depends fundamentally on the person receiving it [10].

When working with tabular data, practitioners typically turn to one of two broad explanation strategies, either feature attribution or rule-based explanations. Feature attribution methods compute importance scores expressing the contribution of each feature to the model's prediction. Rule-based approaches present a combination of logical conditions that must be satisfied for a specific outcome to occur [11]. Anchors [12], for example, derive such if-then rules for individual predictions. Rule-based models are known to be more intuitive and interpretable [8]; however, they do not scale effectively, and rules with many conditions can be difficult to read and understand.

Feature attribution methods, perhaps due to their ability to be used with any model, remain among the most widely adopted explanation strategies. SHAP (SHapley Additive exPlanations) is a popular method that estimates feature importance using Shapley values, a concept derived from cooperative game theory [5]. Similarly, LIME (Local Interpretable Model-Agnostic Explanations) computes feature importance by perturbing input data and fitting a local surrogate model to approximate the behaviour of the original model in the vicinity of the instance being explained [6]. Despite both SHAP and LIME being able to provide valuable insights into model behaviour, their explanations can be difficult for users to interpret, as they may not align well with human mental models. Furthermore, they have been shown to be unstable where similar instances can produce very different explanations [7], and LIME and SHAP explanations can be manipulated by adversarial models [13]. This motivates the need to cross-check and validate explanations from such techniques.

Several factors are important for building trust in explanations, including scalability, human-alignment, stability, and actionability. Scalability means that a method remains useful even when the data or model size becomes large. Human-alignment refers to the extent to which humans agree with the reasons models provide for their predictions [14]. It may be crucial for fostering trust and user acceptance of these systems [15]. Stability requires that similar inputs receive similar explanations [16]. Actionability means that an explanation tells the user what concrete changes can be taken to achieve a different prediction or outcome. With all these aspects in mind, we primarily focus our work on human-alignment, analysing stability across explainers, and supporting feedback-driven refinement.

### *2.2. Visual Analytics for human-aligned explanations*

The growing complexity of ML models has driven the development of visual analytics tools designed to support the interpretation and exploration of model structure and behaviour [17, 18, 19]. Most existing systems are oriented toward helping users inspect models, understand their internal workings, and evaluate performance, rather than toward delivering user-facing explanations directly. Although visual analytics has clear potential to bridge human reasoning and model behaviour, its application to the construction and communication of explanations remains relatively underexplored.

Several rule-based explanation methods use visualisation to facilitate understanding of model behaviour. For instance, RuleMatrix extracts decision rules and organises them into a structured matrix format, enabling interactive exploration of feature relationships and their influence on rule outcomes [20]. Similarly, iForest supports interactive analysis of random forest models by visualising decision paths and predictions [21]. Explainable Matrix represents extracted rules using a matrix-based visualisation framework [22], and VisRuler allows for decision rules to be explored interactively by inspecting the model using various views [23].

Despite the breadth of proposed visualisation and explanation methods, relatively few address the comparison of different approaches. Visagreement [24] offers an interactive visual interface for examining disagreement between explanations, though it is limited to local feature attribution methods. Systems such as ExplainExplore [25] and GALE [26] also support the exploration and comparison of explanations, but likewise focus exclusively on attribution-based approaches.

Current research in human-centred XAI and visual analytics shows the importance for accurate explanations and the importance of presenting these explanations in a way that support human understanding, reasoning, exploration, and decision-making. Previous work has shown that incorporating domain knowledge into model development and explanation generation helps improve human alignment [27]. Building on this perspective, our work introduces a hybrid workflow and interface that integrate rule-based representations, feature attribution methods, and visual analytics to obtain useful insights for feedback-driven refinement of data and models.

### *2.3. The Disagreement Problem and Comparative Visual Analytics*

The limited comparison of the different explanations reflects a broader challenge known as the *disagreement problem*. *Krishna et al.* [4] shows through interviews conducted on practicing data scientists that disagreement between explanation methods is common in real-world workflows, however the users lack streamlined approaches for understanding these differences. Rather than viewing disagreement as a weakness, it is argued that they should be treated as an important source of insight into model behaviour.

This perspective motivates the need for comparative visual analytics that helps users investigate why different explanations agree or disagree. While systems such as Manifold [28] enable interactive comparison of model predictions and feature distributions, Gamut [29] emphasizes that interpretability requirements differ across users and tasks, and Google's What-If Tool [30] supports interactive exploration of model behaviour through counterfactual analysis and data slicing. These systems illustrate the value of comparison and interactive exploration for model understanding, but they do not explicitly support the joint analysis of rule-based and attribution-based explanations generated from the same model.

### *2.4. Evaluating Interpretability*

A recurring theme is that technical faithfulness metrics alone are not enough to establish that an explanation is useful. *Doshi-Velez and Kim* [1] argue that interpretability lacks a rigorous, agreed-upon definition and propose a taxonomy of evaluation approaches: application-grounded, human-grounded, and functionally-grounded, which distinguishes what is being measured and for whom. *Lipton* [31] similarly cautions that interpretability is not a monolithic property: transparency (how a model computes its output) and post-hoc explainability (what can be said about a model post-training) are frequently merged, and optimizing for one does not guarantee the other. *Miller* [32] argues that human explanation is inherently contrastive, selective, and social, and that XAI research often overlooks these properties in favor of complete, model-faithful accounts that are not descriptive to a domain user. *Lage et al.* [33] further show that cognitive complexity is a better predictor of user task performance than model size alone. This reinforces that evaluation must account for how humans process explanations, not just what information is technically present. These perspectives motivate our central claim: that faithfulness and stability, while necessary, do not guarantee that an explanation is human-aligned or actionable, and that human-centered evaluation criteria must be considered alongside technical ones [15, 16]. Our expert evaluation (Section 5) is closest to the application-grounded category: domain experts interpret model explanations on a real dataset.

## 3. Human-Aligned Explanation Framework

In this section, we summarise the XALIGN workflow which is our hybrid visual analytics framework that combines rule-based and attribution-based explanations into an interactive visual interface.

The framework is designed to support explanations that are logical, scalable, human-aligned, stable, and actionable. Satisfying all of these properties simultaneously is a fundamental challenge, as no single explanation approach addresses them adequately in isolation. Rule-based methods express model behaviour as explicit conditions and are therefore well suited to logical, human-interpretable reasoning; however, large rule sets can become difficult to inspect and compare. Feature attribution methods, by contrast, scale well to complex models and high-dimensional data, but their numerical outputs can be difficult to relate to domain knowledge and may vary across methods or local regions of the data.

Table 1 compares feature attribution, rule-based methods, and XALIGN with respect to the factors defined in Section 2. The two explanation types are complementary. Feature attribution scales to complex models, but its numerical output is hard to relate to domain knowledge and varies between methods. Rule-based explanations state conditions explicitly, but large rule sets are hard to inspect and compare. XALIGN shows both in one interface. It does not make a model logical or stable. It allows users to check where these properties hold and where they do not.

### *3.1. Hybrid Visual Analytics Workflow*

Building on this complementarity, we propose a hybrid visual analytics workflow that supports the comparison of multiple explanation methods and the identification of convergence and divergence between the explainers. Figure 1 shows the six steps of the XALIGN workflow. Steps 1 to 3 produce two kinds of evidence for the same model: rules and feature attributions. Step 4 integrates them in one view. In Steps 5 and 6, users explore agreements and contradictions, resolve them using domain knowledge, and refine data, rules, or the choice of explainer. The refinement loop returns to Step 1.

**Step 1: Train a model with a rule representation** The analysis starts with training a predictive model that can eventually be made into a set of decision rules. This may be an inherently rule-based model, such as a decision tree or random forest, or a surrogate model that approximates a more complex classifier. In this work, we use a random forest classifier.

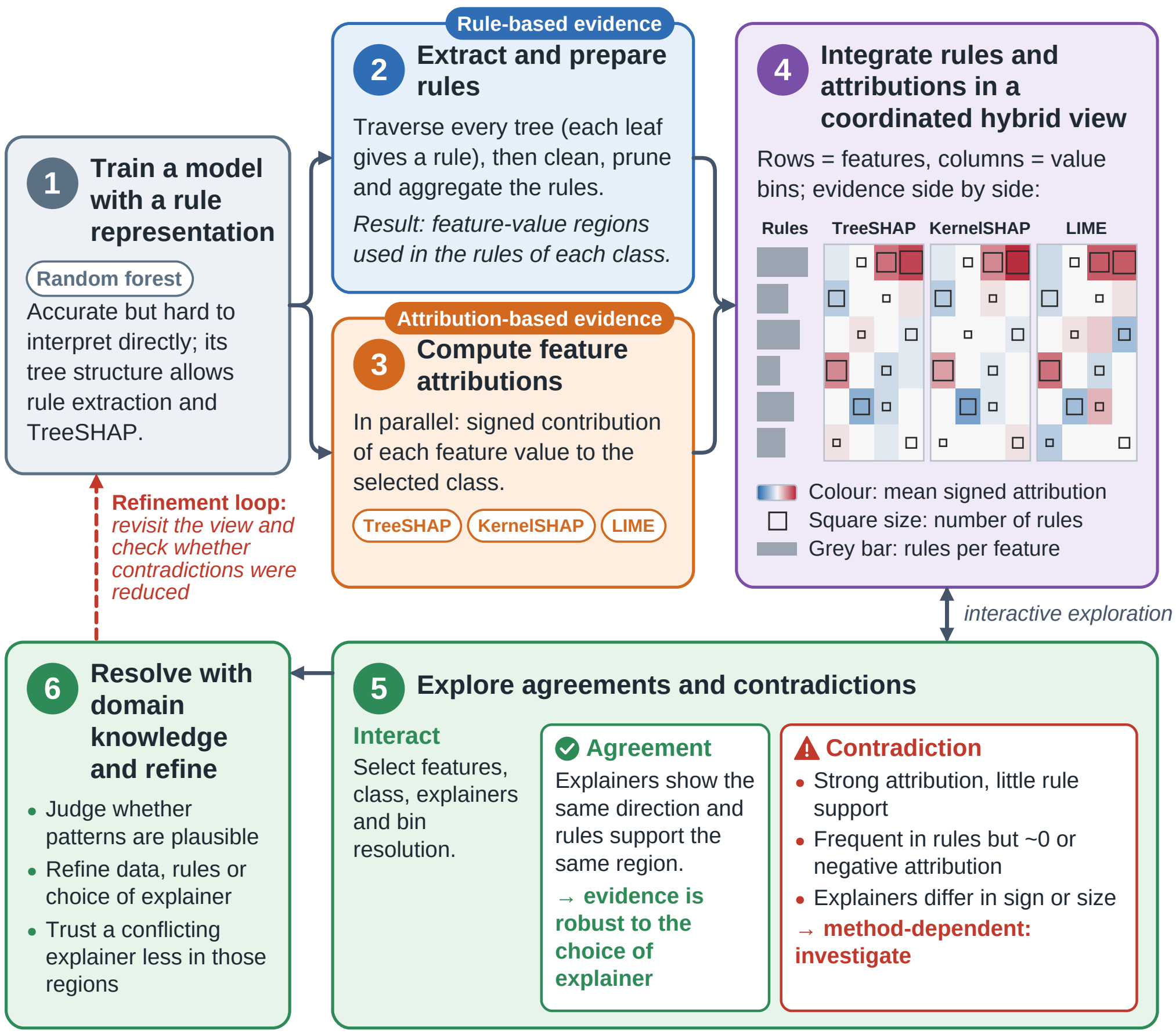

Figure 1: The XALIGN workflow. A model is trained (1). Rules (2) and feature attributions (3) are computed in parallel and integrated in a coordinated view (4). Users explore agreements and contradictions between the explanations (5), resolve them with domain knowledge, and refine data, rules, or explainers (6). The dashed arrow marks the refinement loop.

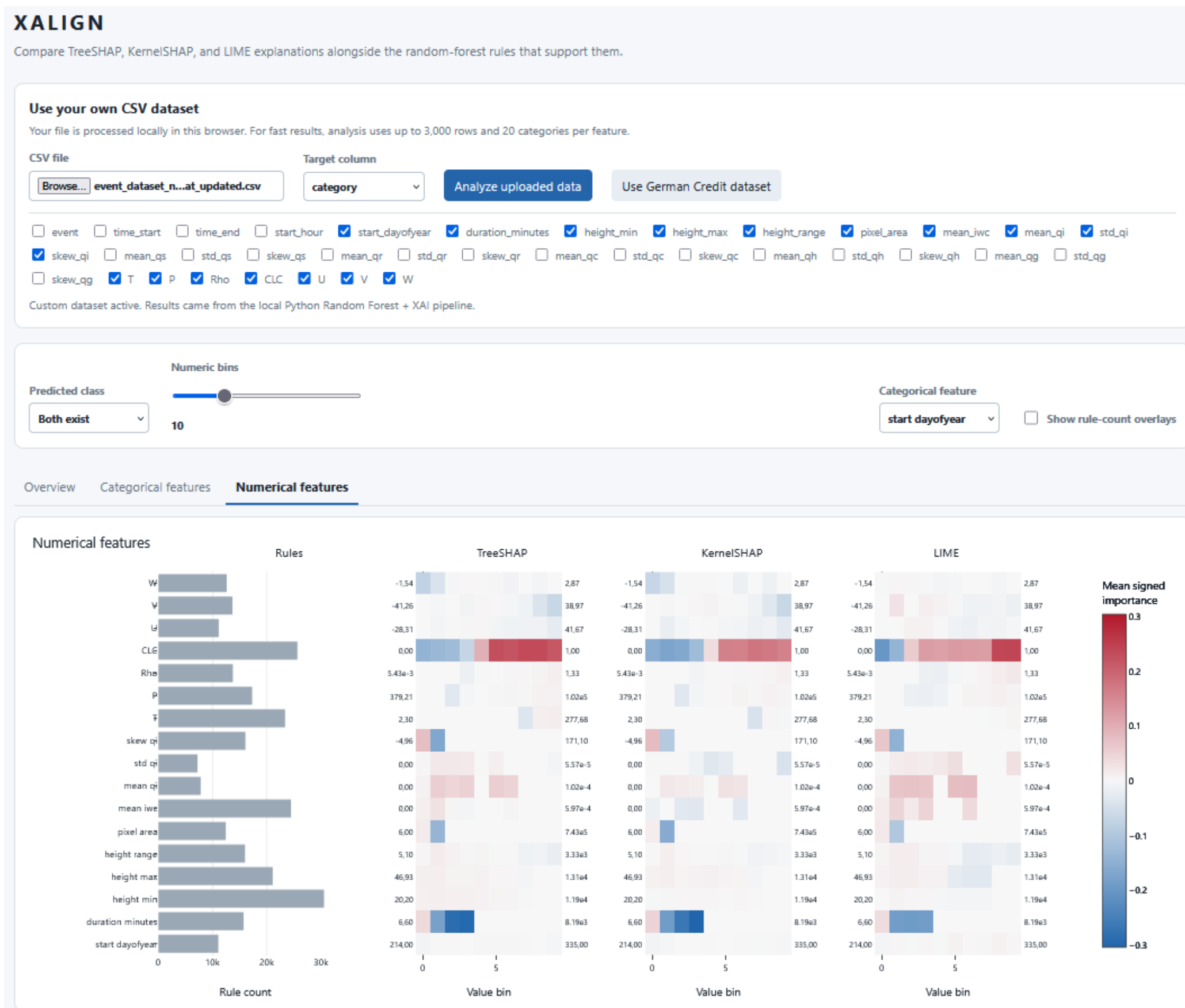

Figure 2: The XALIGN interface (applied to the meteorological dataset). The user can decide the target and which features to include in the analysis. The plots show three different attribution based explainers (TreeSHAP, KernelSHAP, and LIME). This helps to highlight where the methods agree with each other and where they differ. Y-axis: Features, X-axis: Feature bins that can be discretised by a selected value, Colour scale: indicates the feature attribution value. The predicted class and the number of bins can be changed and any features that are not necessary for evaluation can be removed.

Table 1: Comparative analysis of feature attribution methods, rule-based methods, and the proposed hybrid approach across factors relevant to trustworthy explanation.

| Factor | Feature Attribution | Rule-Based Methods | XALIGN |
|---|---|---|---|
| **Logic Type** | Numerical/ Weight-based | Logical (If-Then or numeric ranges) | Multi-modal |
| **Scalability** | High | Low | Limited by rule extraction |
| **Human-Alignment** | Low | High | Checkable against domain knowledge |
| **Stability** | Low | High | Cross-method inconsistencies are visible |
| **Actionability** | Moderate (shows what matters but not how) | High (shows the boundaries for change) | Value ranges and rule boundaries are visible |

Random forests are useful due to their ensemble structure which can achieve strong predictive performance and remain difficult to interpret directly. Their tree-based structure enables rule extraction and supports model-specific attribution methods such as TreeSHAP making it suitable for our analysis

**Step 2: Extract and prepare rules** Each decision tree is traversed path by path, extracting the feature conditions along each branch and pairing them with the corresponding leaf-level prediction. The resulting rules describe the decision logic learned by the ensemble. Because this initial set may be large and redundant, rules are cleaned and aggregated before visual analysis. To do so, along each path, split conditions are combined into a single constraint and repeated splits on the same feature are combined into a single rule. Rules are counted by how many times they appear in a rule leading to a selected class. The resulting representation indicates which feature-value regions occur in rules associated with each predicted class.

**Step 3: Compute feature attributions** Feature attributions are calculated using one or more explanation methods. Although, the framework is not limited to particular attribution techniques, in our case studies we use TreeSHAP [34], KernelSHAP [5], and LIME [6]. TreeSHAP exploits the structure of the trained tree ensemble to calculate SHAP values efficiently. KernelSHAP provides a model-agnostic approximation of Shapley values, while LIME approximates model behaviour locally through a linear surrogate fitted to perturbed samples. Each method produces signed attribution values that indicate the direction and magnitude of a feature's contribution to a selected class prediction. The attributions are computed on the held-out test set. Instances are grouped into their respective feature bins and the mean of the attribution values in that group is used.

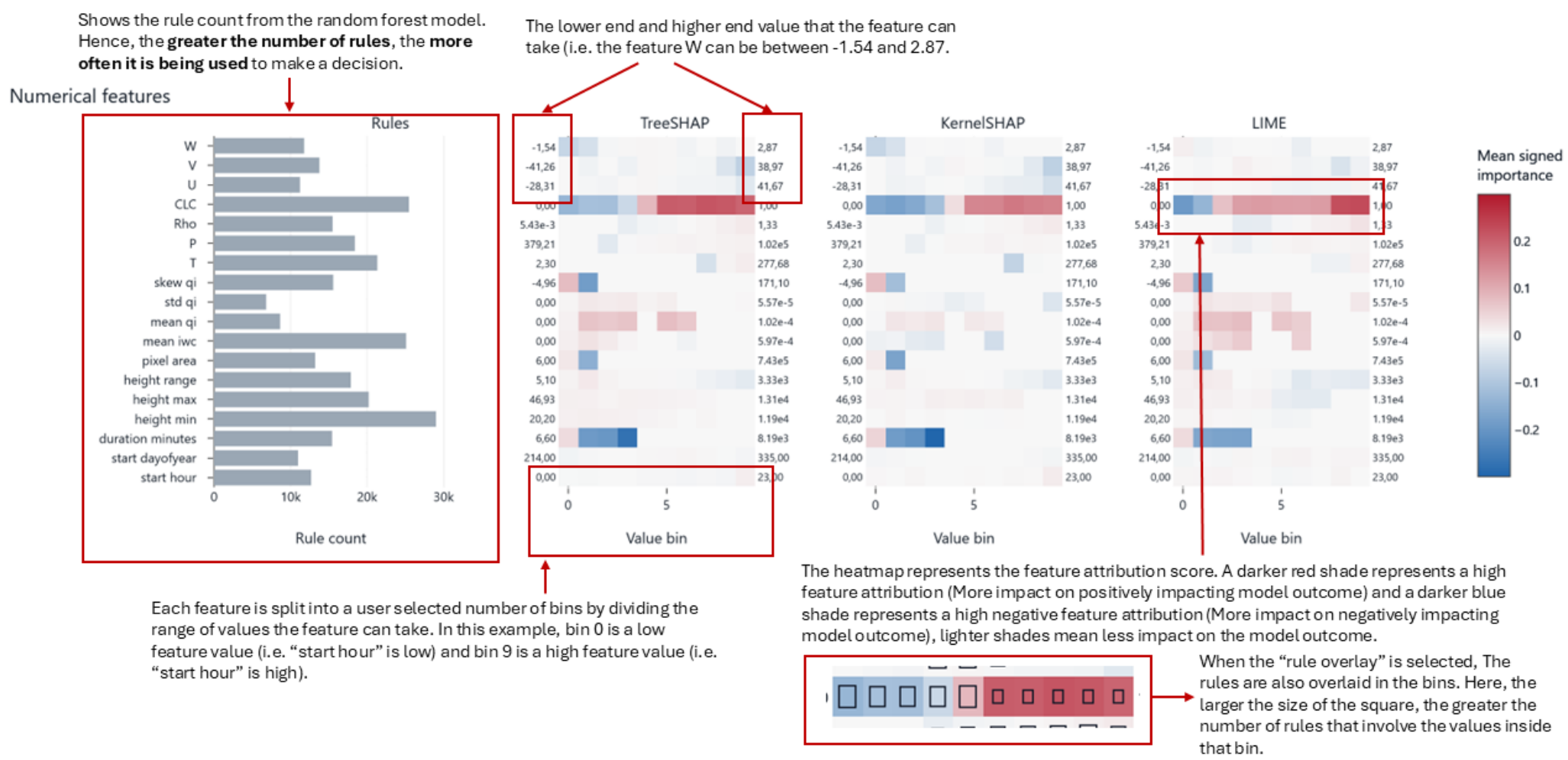

Figure 3: An explanation of how to read the various aspects of the visualisation

**Step 4: Integrate rules and attributions in a coordinated view** The extracted rules and attribution scores are integrated in a hybrid rule-attribution visualization. Figure 2 shows the XALIGN interface on a meteorological dataset, which is explored in more detail in Section 4. Figure 3 explains how to read the view. Features are arranged along the vertical axis, while feature values are discretized into user-defined bins along the horizontal axis. This enables users to inspect how the contribution of a feature changes across its value range. For categorical features, the view represents the presence or absence of a category; for numerical features, it shows the distribution of effects across value bins. The heatmap color encodes the mean signed attribution for each feature-value region. A reversed Red–Blue diverging palette is used. Red indicates that values in that region contribute positively to the selected class, whereas blue indicates a negative contribution; darker shades indicate stronger effects. Squares overlaid on the heatmap encode the number of extracted rules associated with the corresponding feature-value region. Larger squares therefore indicate that the region is more frequently used in the model's rule logic. A rule count does not reflect how many instances pass through the corresponding splits, so a large square does not imply a large effect on the prediction (see Section 4). Grey bars provide an overview of the total number of distinct rules involving each feature. By displaying several attribution methods side by side, the visualisation enables direct comparison between rule-based and attribution-based evidence, as well as between attribution methods.

**Step 5: Explore agreements and contradictions** Users interact with the visualisation by selecting features, classes, attribution methods, and bin resolutions. Agreement is observed when multiple explainers show similar attribution directions and when these patterns are supported by the rule representation. Agreement is judged visually. We do not compute a quantitative agreement measure. For example, a feature-value region with a large rule square and a strong positive attribution across methods indicates that both the extracted decision logic and the attribution methods identify that region as relevant to the selected class. Contradictions become visible when this correspondence breaks down. A feature-value region may receive strong positive attribution but have limited rule support, or it may be frequently present in rules while receiving a negative or near-zero attribution. Attribution methods may also disagree in sign or magnitude. Such patterns are not treated as errors to be removed from the visualisation; instead, they indicate regions where explanations are method-dependent, the model may involve complex interactions, or the available evidence is insufficient for a straightforward interpretation.

**Step 6: Resolve contradictions using domain knowledge and refine the system** The final step of the workflow connects the explanation to the human. Domain knowledge is used to evaluate whether a detected pattern is plausible and whether additional investigation is required. Where necessary, decision about the refinement of the data, rules, or the choice of explainer can be made. If a particular explainer consistently conflicts with both domain knowledge and the other methods, the user may decide to further investigate these regions. The workflow also supports feedback-driven refinement. Once the domain user has identified data segments that need further evaluation, they can propose changes in the data or model as required. After changes are made, the visualisation can be revisited to assess whether contradictions have been reduced or whether new inconsistencies have emerged. Thus, the framework treats explanation as an iterative human-AI process rather than a final model output. The case studies in Section 4 illustrate Steps 1 to 5. Applying the refinement of Step 6 is left as future work.

### *3.2. Supporting Human-Aligned and Trustworthy Explanations*

The framework provides users with the information needed better rely on the model. When the methods agree, it can increase the confidence that a feature-value pattern occurs regardless of the explainer used. When the methods do not agree, it exposes uncertainty and suggests to the user that they should be more careful in relying on the explanation.

It is important to note that agreement does not mean that the model is correct, fair, or causally valid. It only shows that the different explanation methods provide compatible descriptions of the model's learned behaviour. Allowing the user to see the agreements and disagreements, the framework helps to shifts explainability towards human-centric assessment from the user. So, rather than trusting a single explanation, it helps the user determine when an explanation is sufficiently supported and when it should be further investigated.

## 4. Case Studies

To illustrate the use of the tool, we present two case studies from two different application domains. The first case study looks at a widely used dataset in Machine Learning research and is publicly available. The second is a more specialised case study where we use a real-world meteorology dataset.

### *4.1. German Credit*

The German Credit dataset [35] contains 1,000 credit applicants described by 20 attributes. Seven attributes are numerical: age, duration in months, credit amount, installment rate, present residence since, number of existing credits at this bank, and number of people liable to provide maintenance. The others are categorical, for example purpose of the loan, employment status, and housing. Each applicant is labelled as a good or bad credit risk. We train a Random Forest with 200 trees and obtain 75.67% Accuracy (F1 Score = 0.840) on 300 records (test data). All views show attributions for the class bad credit risk. Positive values push the prediction towards this class. The mixed feature types make the dataset a useful test for the rule-attribution view.

The German Credit dataset has become a standard benchmark for evaluating classification algorithms, particularly in contexts requiring model interpretability and fairness, making it especially relevant for testing explainability methods like SHAP and LIME. Its combination of mixed feature types, class imbalance, and real-world financial implications makes it a suitable testbed for assessing how different explanation techniques handle complex, consequential decision-making scenarios and was hence used in this work to demonstrate the utility of the proposed interactive tool.

The rule-attribution visualisation for the German Credit Dataset is presented in Figures 5 and 4 (Numerical data and categorical data). The categorical views show several regions of strong cross-method agreement. For status of existing checking account, the clearest pattern occurs for the categories “no checking account” and “< 0 DM”. For "no checking account", TreeSHAP and LIME indicate a positive contribution when these category indicators are absent (state 0) and a negative contribution when they are present (state 1), with particularly strong negative attribution from LIME for “no checking account”. KernelSHAP shows a weak positive contribution for state 0 and is close to neutral for state 1. The rules also support this where the larger box for state 0 also suggests positive contribution. For “<

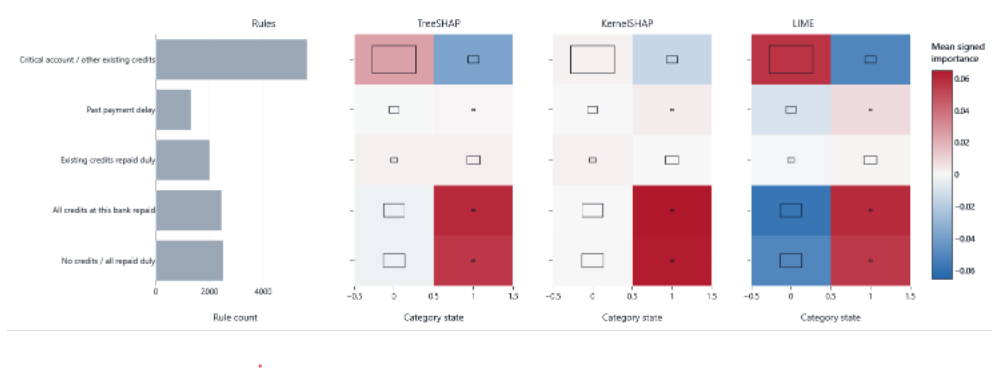

(a) Credit History

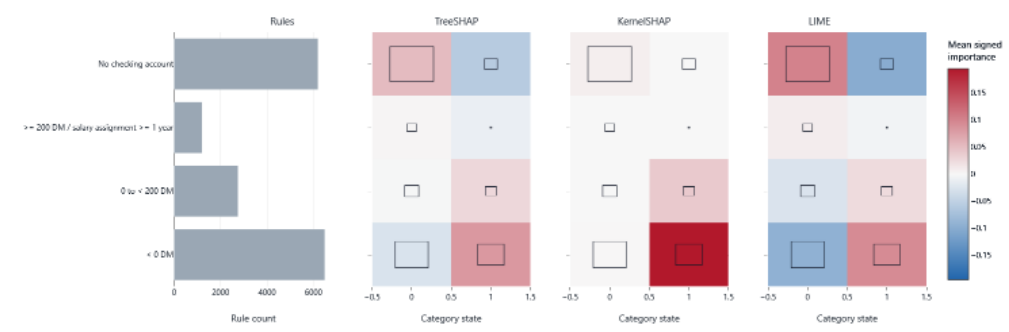

(b) Status of existing checking account

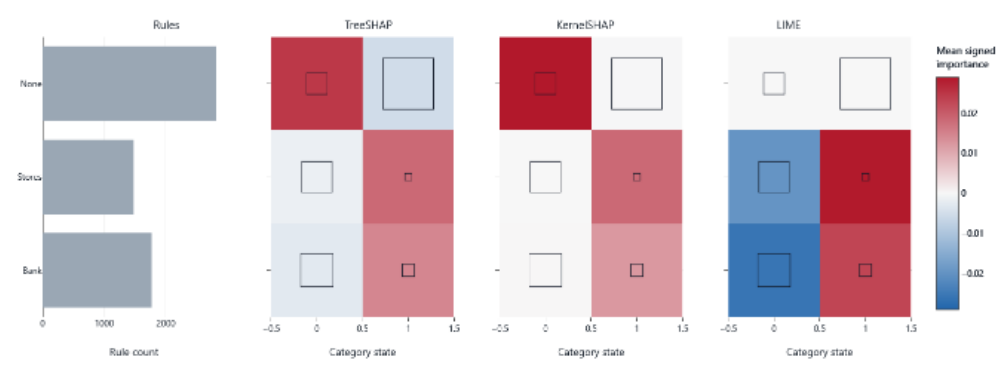

(c) Other Installment Plans

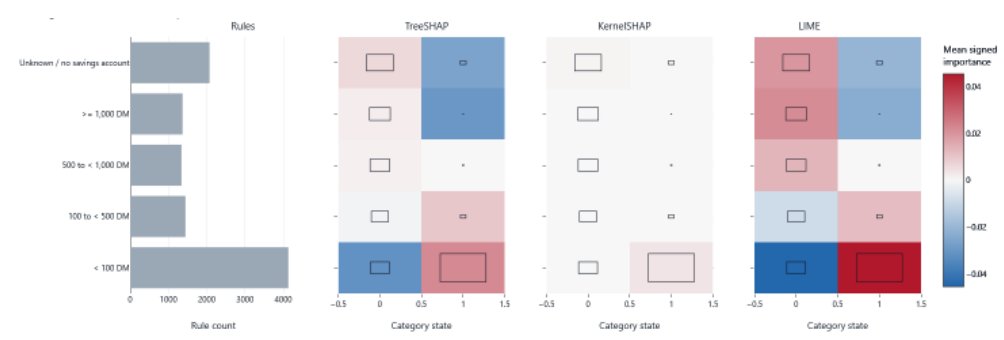

(d) Savings account bonds

Figure 4: Each categorical feature can be investigated individually. Here the Rule-attribution visualsation is applied on four features. The class is bad credit risk.

0 DM", there is a suggestion by the three attribution methods that the presence of this category is a positive indicator of "bad" credit, however, the rule based approach suggests no strong preference and hence the impact of there being "< 0 DM" present is something the user should take with more caution.

A clear agreement between methods is found for savings account/bonds, especially for "< 100 DM." All explainers assign a negative contribution to state 0 and a positive contribution to state 1, while the latter state also has substantial rule support. This is one of the most stable categorical patterns in the views. The evidence is less consistent for the other savings categories. For example, TreeSHAP and LIME both assign negative attribution to the presence of "unknown/no savings account" and ">= 1,000 DM," whereas KernelSHAP is close to neutral. These differences are not direct sign reversals, but they show that the estimated strength of these categories depends on the explainer.

For credit history, the category "critical account/other existing credits" shows broad directional agreement: TreeSHAP and LIME attribute a positive contribution to state 0 and a negative contribution to state 1, while KernelSHAP shows the same tendency at lower magnitude. The categories representing fully repaid credit histories also show a consistent positive attribution when present, especially for "all credits at this bank repaid" and "no credits/all repaid duly". However, the rule support for state 1 is very small, so the rules do not corroborate this pattern. The categories "past payment delay" and "existing credits repaid duly" have few rules and near-zero attributions. We treat them as low-evidence regions, not as disagreement. "Other installment plans" is a contradiction. All three attribution methods agree with each other, while the rules point in the other direction. This feature should be interpreted with caution.

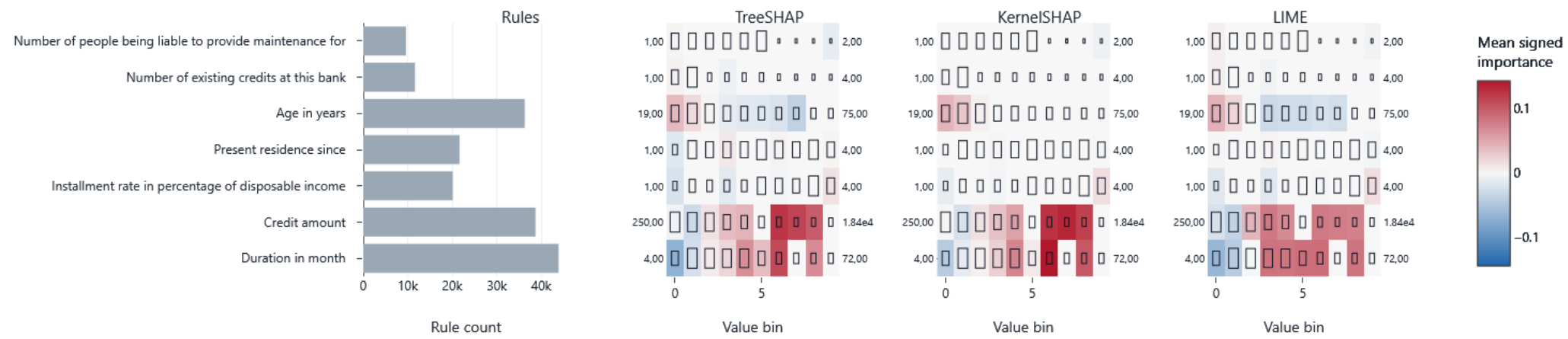

Figure 5: Numerical Features can be investigated individually. Here the numerical features from the German Credit Dataset are shown for the three explainer types. The class is bad credit risk.

The numerical-feature view shows the importance of the features "duration in month" and "credit amount". These features have the largest overall rule counts and the most pronounced attribution patterns. Across TreeSHAP, KernelSHAP, and LIME, low duration and low-to-moderate credit-amount bins tend to have negative or neutral contributions, whereas several higher bins have positive contributions. The strongest positive regions occur at high duration bins and at high credit-amount bins, and these regions are generally supported by larger rule boxes. Although the exact peak bins vary between explainers where for example, KernelSHAP concentrates a particularly strong positive effect in selected high-value bins while LIME distributes positive contributions more broadly, the overall direction and non-linear threshold-like pattern are consistent.

Age provides a useful example of agreement that is incomplete. TreeSHAP and LIME both attribute positive effects to the youngest age bins and negative effects to several middle-to-higher bins. KernelSHAP agrees on the positive contribution at the youngest bins but is much closer to neutral elsewhere. Thus, the direction identified by TreeSHAP and LIME is not fully corroborated by KernelSHAP, suggesting that age-related effects are weaker or more sensitive to the attribution method than the effects of duration and credit amount. The remaining numeric variables—present residence, installment rate, number of existing credits, and number of people liable for maintenance—show comparatively small and scattered attributions. Even where rule counts are non-zero, the lack of a stable high-magnitude signal across explainers suggests that these features should not be given the same interpretive weight.

Overall, the strongest warranted conclusions are those supported simultaneously by large rule boxes and consistent attribution signs, notably the "no checking account", the "< 100 DM" savings category, loan duration, and credit amount. Regions with weak, neutral, or explainer-dependent attributions should be presented as areas of interpretive uncertainty rather than forced into a single explanation. This is the role of the hybrid visualisation: agreement provides cross-method evidence for a model pattern, while disagreement highlights feature-value regions where reliance on any single explainer should be tempered.

This case study supports our broader argument: the value of the hybrid rule–attribution visualisation lies in showing specific disagreements between explanation methods and in indicating how much trust to place in different regions of the explanation space.

### *4.2. Meteorology Dataset*

The second case study investigates meteorological data consisting of five months of high resolution cloudnet observation data [36] and Climate prediction ICON Model data [37]. The dataset used for this task consists of ice water content (`iwc`) values for time-height from the period from the 1st August 2020 to 31st December 2020 and for the heights between 0 and 12,000 meters above sea level.

The data was processed by aligning the iwc data from the model and observation grids. The observed ice water content was subtracted from the model ice water content, and thresholding was used to categorise the results into where there was a reading present for both model and observation, where there was only a reading in the model data, and where there was only a reading in the observation data. With these categories, the data was structured into three classes: "Both exist": Where there was a reading present for both model and observation data, "Model Only": Where only Model data existed, and "Observation Only": Where there was only observation data.

To create events from this data, bounding boxes were derived around each category using a temporal abstraction strategy [38] where a moving window along the time-height data was categorised into the selected classes based on the summarized values. A height restriction was placed on the events to disallow events that were too large. This gave us the data required for the machine learning task. The prediction task was defined as the identification of disagreement between the model and observational data. For this, a multi-class supervised learning task was defined. The dataset consists of 2148 instances with three classes and 32 attributes. In this use-case, all attributes are numerical. The attributes consist of a variety of physical and geometric features including:

**Geometric and Temporal Descriptors of the bounding boxes:** Minimum height (`height_min`), maximum height (`height_max`), vertical range (`height_range`), pixel area (`pixel_area`), event duration (`duration_minutes`), start hour of day (`start_hour`), and day of year (`start_dayofyear`).

**Physical Features:** Temperature (`T`), Pressure (`P`), Density (`Rho`), Cloud Cover (`CLC`), Zonal Wind (`U`), Meridional Wind (`V`), Vertical Wind (`W`)

**Model Parameters:** Mean, Standard Deviation, and Skew of Cloud liquid water content (`QC`), cloud ice content (`QI`), rain mixing ratio (`QR`), hail mixing ratio (`QH`), Graupel mixing ratio (`QG`), Snow mixing ratio (`QS`)

A random forest model consisting of 100 decision trees was used to classify the three classes. A stratified 4-fold cross validation was undertaken while preserving class proportions across each validation split and the XALIGN workflow was then undertaken on this dataset.

In Figure 6, it can be observed that events in which both the model and observations indicate the presence of ice water content are more strongly associated with higher cloud cover (`CLC`), lower `QI` skewness, very short event durations, and mid-to-high temperatures. In contrast, Model Only events are characterised by lower cloud cover, lower temperatures, large vertical extents (low minimum and high maximum heights), and extreme pressure conditions, with both very low and very high pressure being associated with this category. The Observation Only events exhibit some similarities with the Model Only category, particularly in their association with lower cloud cover, but are distinguished by higher temperatures and intermediate `QI` skewness. Overall, the rule attribution visualisation suggests that cloud cover, temperature, and the distribution of ice water content are important factors differentiating model–observation agreement from cases where ice water content is detected exclusively by either the model or the observations. The contrasting temperature patterns between Model Only and Observation Only events further indicate that the model's representation of ice water content may differ from observations under different atmospheric conditions. In this case of the meteorology data, there is very good agreement between the three attribution techniques with only the intensity changing while the general pattern remains the same.

When we look at the rules overlaid onto the attributions, in this meteorology case, the match or mismatch is less clear than for the German Credit Dataset. The meteorology case is a three-class problem, whereas the German Credit case is a binary problem. When we look at the rule count for the feature `CLC`, we can see that for the class "Both Exists", the rule count is high across the bins although the attribution score shows a clear difference between bins. However, the difference between rules is more clear in the other two classes where there are many more rules in the lower bins. A feature can be used early in a tree, splitting a large portion of the samples, or deep in a tree, splitting very few. Both situations generate rules and inflate the square size. But attribution methods weight by the actual change in prediction probability, so a deep, rarely-triggered split contributes little to mean attribution even though it exists in many rule paths. This can also be seen in the feature `height_min` where it consistently has a large rule count across the three classes, although the heatmap is mostly pale. This means the feature is frequently used as a splitting variable and hence the model is very dependent on the existence of this feature, although it does not strongly direct the prediction to a single class.

### *4.3. Discussion*

The case studies show that agreement is most informative when it occurs across the explanation methods. In the German Credit data, duration, credit amount, checking-account status, and selected savings-account categories exhibited similarly coherent patterns. Although this convergence does not prove that the learned relationship is correct or causal, it provides stronger evidence that the explanation is not an artifact of one particular technique. Hence, agreement serves as a practical indicator of explanation robustness and can help users identify model behaviour with greater confidence.

The framework also makes contradictions visible. TreeSHAP, KernelSHAP, and LIME do not always assign the same direction or magnitude of importance to the same feature-value region. These differences may result from the distinct assumptions underlying the methods: TreeSHAP exploits the structure of the tree model, KernelSHAP approximates Shapley values through sampling, and LIME uses local linear surrogate models. Disagreement may therefore arise in regions with non-linear decision boundaries, feature interactions, limited data support, or locally unstable behaviour. Rather than treating these discrepancies as noise to be hidden, the proposed visualisation treats them as useful diagnostic signals. They identify regions in which an analyst should investigate more carefully before accepting an explanation or acting on a prediction.

These case studies also show that both the rules and the attributions provide complementary but different perspectives about the model. The rule count shows how much the model depends on the feature to partition the classes whereas the attributions show how much the feature moves the prediction towards or against a particular class. So, although high rule count is also a form of importance, the underlying meaning can be different to the attributions which is important to consider. This dual aspect provides the user with various ways of investigating the model further. Also, the meteorological problem is three-class, requiring a separate visualisation per class with the sign of attributions relative to each specific class. The German credit problem is binary, so each visualisation has an unambiguous positive direction where positive attribution means pushes toward one class, negative toward the other. This makes the German credit plots easier to interpret intuitively meaning the complexity of the model will also impact the ability to obtain useful insights into the model behaviour.

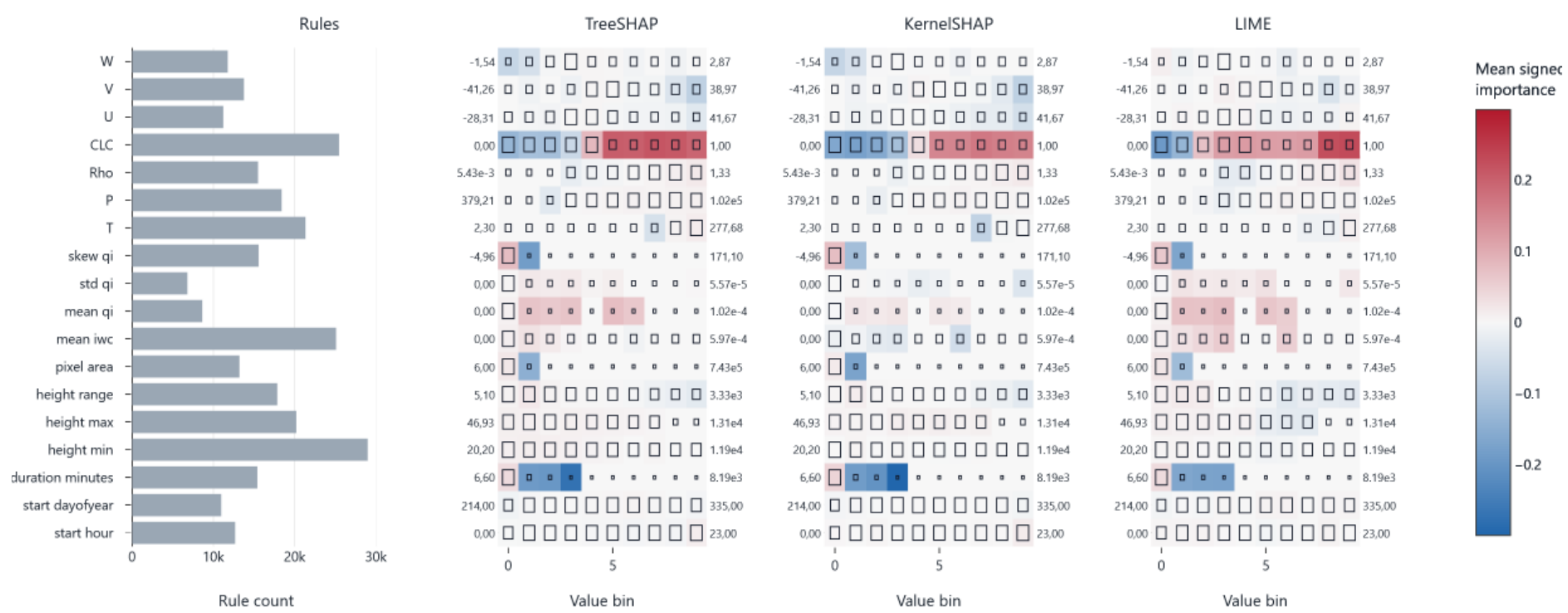

(a) Both Exist

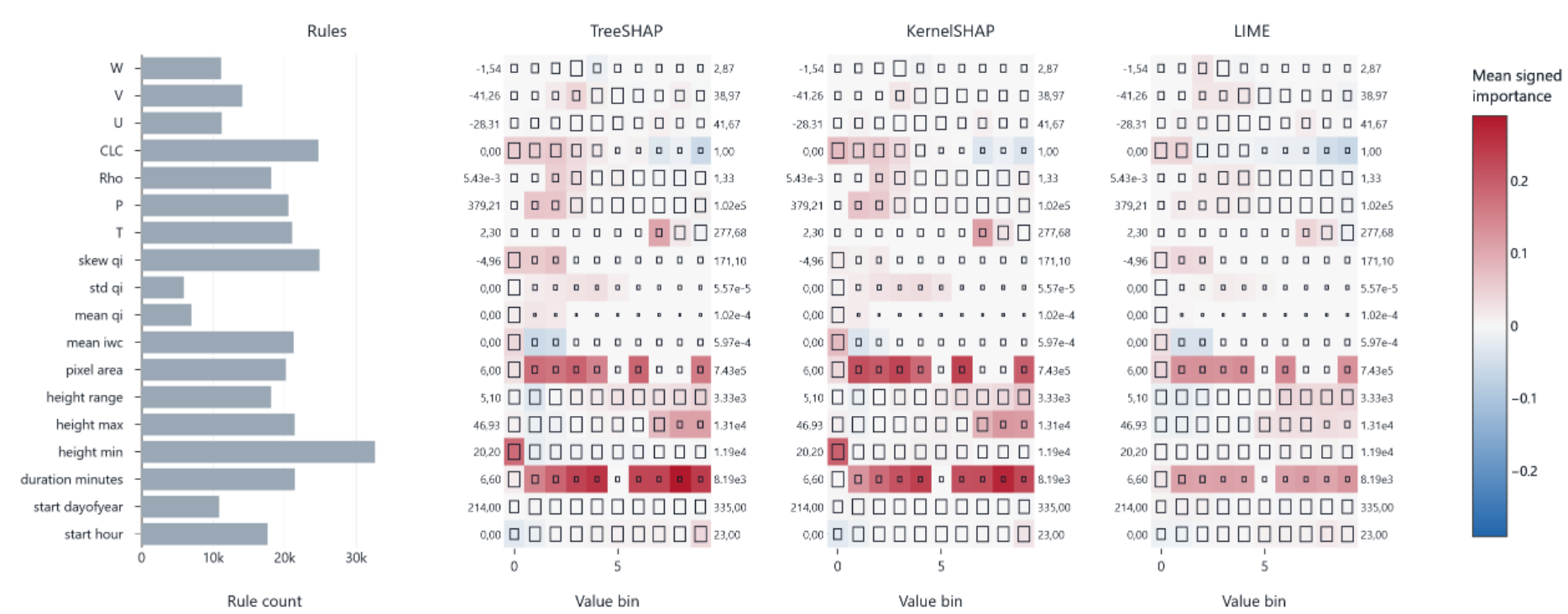

(b) Model Only

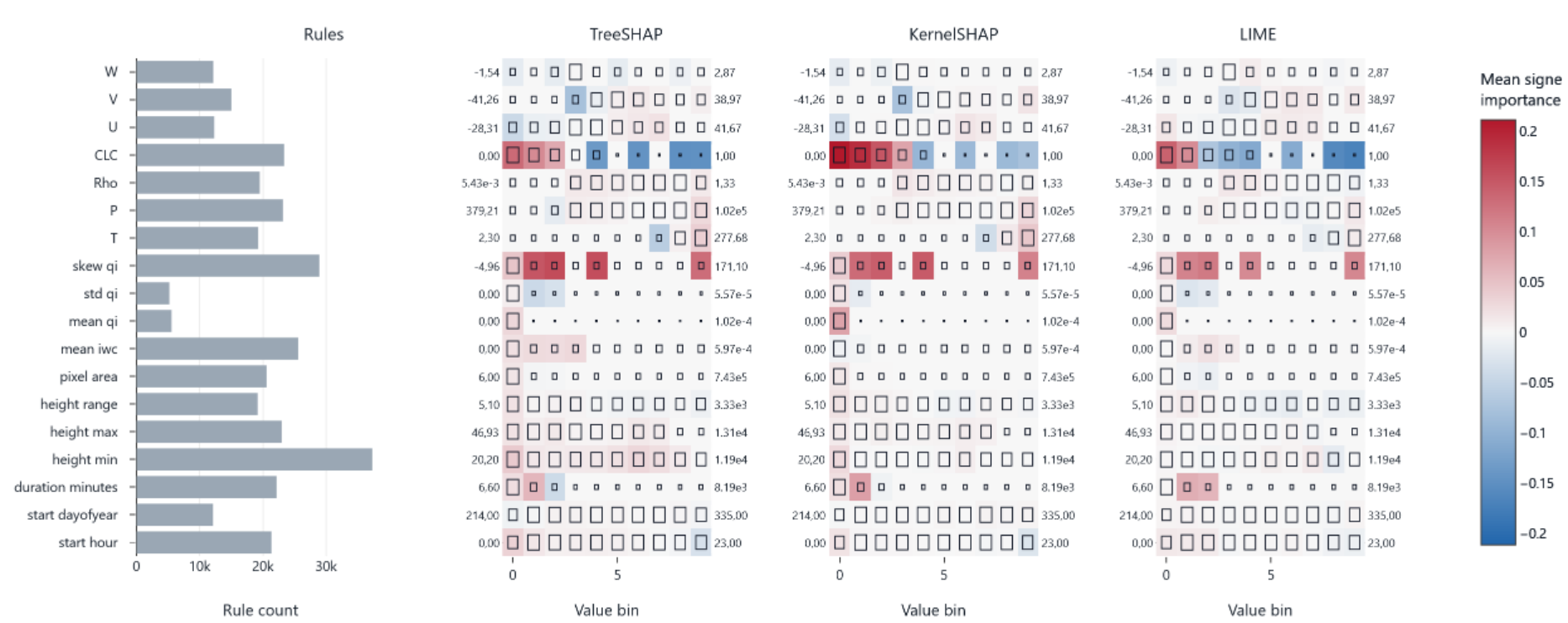

(c) Observation Only

Figure 6: The XALIGN rule visualisation for the meteorological use-case using a reduced feature subset for the three classes (a) Both model and observation data are present, (b) Only Model data is present, and (c) Only Observation data is present.

## 5. Expert Evaluation

We conducted a qualitative formative expert evaluation to assess whether the XALIGN tool and visualisation support expert users in understanding model predictions, identifying influential meteorological features, and forming calibrated judgements about model reliability. Given the specialised nature of the tool, we recruited a small panel of three XAI and/or meteorology experts rather than a large convenience sample, following established practice for expert evaluations of analytical visualisation systems. Each session was a single semi-structured think-aloud interview, so the study provides initial feedback and not evidence of long-term use. Long-term case studies with domain experts, as proposed in the MILCs framework [39], are a natural next step.

### *5.1. Study Design and Participants*

The participants of the evaluation consisted of:

- P1, a visualisation expert with in-depth knowledge of feature attribution and rule-based models.
- P2, a visualisation expert working regularly with feature attribution visualisations.
- P3, a meteorological data expert working regularly using ICON weather prediction model data.

Participants were recruited to reflect the two primary user groups the tool is designed for, XAI practitioners seeking to audit model behaviour, and domain scientists seeking to interpret model predictions.

Each session was conducted individually as a semi-structured think-aloud interview. Prior to the evaluation tasks, participants received a brief introduction and a walkthrough of the tool's functionality and instruction on how to read the visualisations. This introduction was done using the German Credit Dataset. Following this, an introduction to the meteorology data and the corresponding Machine Learning task was provided to the expert. Once the expert was satisfied that they understood the tool and the dataset, we moved to the formal evaluation. Sessions proceeded through four phases. In the first phase, participants were shown the visualisation for the model only class and asked a series of structured interpretation tasks to confirm their understanding of the visualisation. In the second phase, participants were asked open-ended questions to see if they were able to gain insights from the tool. The third phase was an opportunity for the expert to rate using a five Likert-scale measure, statements covering their understanding of model reasoning, feature identification, class differentiation, resistance to over-trust, and anticipated practical use. The fourth and final phase comprised a structured debrief with open-ended questions aiming to get feedback and use cases.

### *5.2. Evaluation Results*

Overall, participants responded positively to the XALIGN visualisation, identifying several strengths while also highlighting areas for improvement. In the Likert scale rating questions, all of the participants strongly agreed that they would use this visualisation for other datasets. They also all agreed that they could identify which features were driving the prediction and that the visualisation would help them to not over trust a model. However the scores for "I understand why the model made this prediction" were varied with one participant stating they agreed and the other two were unsure. The results are discussed further below in terms of usability and utility, interpretation challenges, and design feedback.

#### *Usability and Utility*

All three participants found the tool useful for identifying features that may be driving model predictions, and reported that it was easy to see at a glance which features carried the most weight for a given class. The visualisation was perceived as compact and uncluttered, with participants noting that it did not introduce unnecessary visual noise. All participants indicated that they had other datasets they would be interested in applying the tool to, suggesting that they perceived it as practically transferable beyond the specific meteorological context. Participants also broadly agreed that the visualisation would help them avoid over-trusting a model prediction, supporting one of the core aims of the tool.

#### *Interpretability*

Overall, the feature attribution panel and its relationship to feature value ranges were found to be relatively straightforward to interpret. However, the connection between the attribution results and the extracted rules was more challenging. This challenge appeared to be tied to familiarity with the underlying explanation methods where the participants with a background in XAI were able to relate the two representations more readily, whereas the meteorology data expert who had less knowledge of XAI found this link less intuitive. This suggests a steeper learning curve for users who are less familiar with how rule-based models are constructed and how rules relate to attribution scores.

All the participants noted that while the tool effectively highlights which features are important, it is harder to derive a clear understanding of *why* a particular prediction was made indicating the impact due to the interaction between features remains a challenge. The experts also suggested that a natural language summary accompanying the visualisation could help bridge this gap for users without an XAI background. Additionally, two of the participants were confident that given more time to familiarise themselves with the tool, they believed they would be able to extract more meaningful insights which shows that, like with many expert tools, there is an inherent learning curve that may diminish with continued use.

*Design Feedback and Post-Study Refinements*

A recurring suggestion from participants was the ability to compare the three predicted classes side by side, rather than examining each class in isolation. Based on this feedback, an additional tab was incorporated into the tool after the user study, allowing users to view either the three attribution methods for a selected class in parallel, or the same attribution method across all three classes simultaneously. This extension is intended to address the expressed need for cross-class comparison and to reduce the cognitive overhead of switching between views. It has not been evaluated.

### *5.3. Limitations*

This evaluation was preliminary and provided us with feedback for refinement and a further larger study. Limitations of the study include the small number of participants and no other comparison to XALIGN. The tab for cross-class comparison was added after the study and has not been evaluated. Also, only one of our participants was familiar with the domain data and hence to gain more insights into how this would help in particular domains such as meteorology, a more focused study where this tool is presented to meteorologists is required.

## 6. Conclusions

In this paper, we argue that relying on a single XAI approach is insufficient for trustworthy, human-aligned AI systems. To address this, we introduce a hybrid approach that combines rule-based explanations with feature attribution methods, presented through an interactive visual analytics interface for unified comparison and evaluation. Case studies on the German Credit dataset and a meteorology dataset illustrated that where feature influences were strong, explanation methods converged reliably. Conversely, for features with weak or scattered effects, methods diverged, increasing the risk of misinterpreting model behaviour. The interactive visualisation approach provides several practical benefits:

- Agreements and discrepancies between explanation methods are shown simultaneously, so users do not need to compare separate outputs manually.
- Users can drill down into specific features and bins and cross-reference against domain knowledge.
- Contradictions that would likely go unnoticed in isolation are made visible and become meaningful points for investigation.

These results indicate the importance of integrating interactive visual analytics into explanation validation processes. Furthermore, an expert evaluation found XALIGN useful for identifying important features. In this small evaluation, the results suggest that the XALIGN visualisation is most immediately suited to users with at least a basic understanding of feature attribution and rule-based methods. For domain scientists without this background, additional scaffolding such as natural language summaries or guided walkthrough may be needed to support effective interpretation. Hence a useful avenue for future work could involve the use of LLMs to provide natural language summaries of the insights. This would also eventually allow the user to automatically flag contradictions for expert review which would further enhance the feedback loop between human insight and visualisation. Future work also includes support for feature interactions, which participants identified as a gap, and a study of the refinement loop (Step 6).

## Acknowledgements

This work was supported by the German Federal Ministry of Research, Technology and Space within the Project WarmWorld (01LK2204A), the Deutsche Forschungsgemeinschaft (DFG, German Research Foundation) under Germany's Excellence Strategy EXC 3037 – 533607693 – Our Dynamic Universe" and the Federal Ministry of Education and Research of Germany and the state of North-Rhine Westphalia as part of the Lamarr Institute for Machine Learning and Artificial Intelligence.